\documentclass[10pt,twocolumn,letterpaper]{article}

\usepackage{cvpr}              

\usepackage{times}
\usepackage{epsfig}
\usepackage{graphicx}
\usepackage{amsmath}
\usepackage{amssymb}
\usepackage{enumitem}
\usepackage{amsbsy}

\usepackage{booktabs}
\usepackage{siunitx}
\usepackage{multirow}
\usepackage{epstopdf}

\usepackage{xcolor}
\usepackage[font=small,labelfont=bf]{caption}  
\usepackage{makecell}
\usepackage{tabulary,colortbl}

\definecolor{LightCyan}{rgb}{0.9059,0.9961,1}

\usepackage[pagebackref,breaklinks,colorlinks]{hyperref}

\usepackage{xcolor}
\usepackage[font=small,labelfont=bf]{caption}  
\usepackage{makecell}
\usepackage{tabulary}
\definecolor{demphcolor}{RGB}{144,144,144}

\definecolor{mygray}{gray}{0.4}
\usepackage{pifont}
\newlength\savewidth

\newcommand{\tablestyle}[2]{\setlength{\tabcolsep}{#1}\renewcommand{\arraystretch}{#2}\centering\footnotesize}
\makeatletter\renewcommand\paragraph{\@startsection{paragraph}{4}{\z@}
  {.5em \@plus1ex \@minus.2ex}{-.5em}{\normalfont\normalsize\bfseries}}\makeatother
\newcolumntype{C}[1]{>{\centering\arraybackslash}p{#1}}
\newcolumntype{R}[1]{>{\raggedleft\arraybackslash}p{#1}}
\newcolumntype{L}[1]{>{\raggedright\arraybackslash}p{#1}}

\newcommand{\specialcelll}[2][l]{%
  \begin{tabular}[#1]{@{}l@{}}#2\end{tabular}}

\definecolor{asparagus}{rgb}{0.53, 0.66, 0.42}

\def\cvprPaperID{10711} 
\def\confName{CVPR}
\def\confYear{2022}

\begin{document}

\title{MLP Architectures for Vision-and-Language Modeling: An Empirical Study}

\author{Yixin Nie\thanks{\,\, Equal contribution.}\,\,$^1$, Linjie Li\footnotemark[1]\,\,$^2$, Zhe Gan$^2$, Shuohang Wang$^2$,\\ Chenguang Zhu$^2$, Michael Zeng$^2$, Zicheng Liu$^2$, Mohit Bansal$^1$, Lijuan Wang$^2$\vspace{5pt}\\ 
  $^1$UNC Chapel Hill \quad $^2$Microsoft Corporation\\
  \texttt{\small\{yixin1, mbansal\}@cs.unc.edu} \\
  \texttt{\small\{lindsey.li, zhe.gan, shuowa, chezhu, nzeng, zliu,  lijuanw\}@microsoft.com}
  }

\maketitle

\begin{abstract}
  We initiate the first empirical study on the use of MLP architectures for vision-and-language (VL) fusion. Through extensive experiments on 5 VL tasks and 5 robust VQA benchmarks, we find that: ($i$) Without pre-training, using MLPs for multimodal fusion has a noticeable performance gap compared to transformers; ($ii$) However, VL pre-training can help close the performance gap; ($iii$) Instead of heavy multi-head attention, adding tiny one-head attention to MLPs is sufficient to achieve comparable performance to transformers. Moreover, we also find that the performance gap between MLPs and transformers is not widened when being evaluated on the ``harder" robust VQA benchmarks, suggesting using MLPs for VL fusion can generalize roughly to a similar degree as using transformers.
  These results hint that MLPs can effectively learn to align vision and text features extracted from lower-level encoders without heavy reliance on self-attention. Based on this, we ask an even bolder question: can we have an all-MLP architecture for VL modeling, where both VL fusion and the vision encoder are replaced with MLPs?
  Our result shows that an all-MLP VL model is sub-optimal compared to state-of-the-art full-featured VL models when both of them get pre-trained. However, pre-training an all-MLP can surprisingly achieve a better average score than full-featured transformer models without pre-training. This indicates the potential of large-scale pre-training of MLP-like architectures for VL modeling and inspires the future research direction on simplifying well-established VL modeling with less inductive design bias. Our code is publicly available at: \url{https://github.com/easonnie/mlp-vil}.
\end{abstract}

\vspace{-3mm}
\section{Introduction}
Transformers~\cite{vaswani2017attention} have greatly advanced the state of the art in natural language processing~\cite{kenton2019bert}, vision-and-language modeling~\cite{tan2019lxmert, li2019visualbert, lu2019vilbert, chen2020uniter}, and more recently computer vision~\cite{dosovitskiy2020image,liu2021Swin}.
The success of transformers is partially due to its simplicity, which makes it possible to leverage large-scale training data and distributed computational power at ease~\cite{zhai2021scaling, brown2020language}. It also reveals the evolutionary paradigm shift of neural network design from customized models with inductive biases to unified layers with as few architecture priors as possible. To push this trend further, more recent works~\cite{tolstikhin2021mlp,touvron2021resmlp,liu2021pay} have simplified the transformer by replacing the self-attention block with a token-wise feed-forward block. 
This leads to an architecture containing only multi-layer perceptrons (MLPs), reminiscing of the most primitive form of neural networks.

The recurring interest in all-MLP architectures is theoretically hinted by the idea that MLPs can approximate arbitrary functions to arbitrary precision~\cite{hornik1989multilayer}.
In transformers, the attention mechanism introduces the inductive bias that token-wise interactions should be \emph{dynamically} parameterized based on the input representations \cite{vaswani2017attention}.
Therefore, one can remove attention to verify whether such inductive bias could be acquired with \emph{statically} parameterized MLPs.
Empirically, the concept seems to have been proven, since newly proposed all-MLP architectures have achieved reasonably good performance for image recognition~\cite{tolstikhin2021mlp, touvron2021resmlp} and masked language modeling~\cite{liu2021pay}.

\begin{figure*}[t!]
	\centering
    \includegraphics[clip,width=0.90\textwidth]{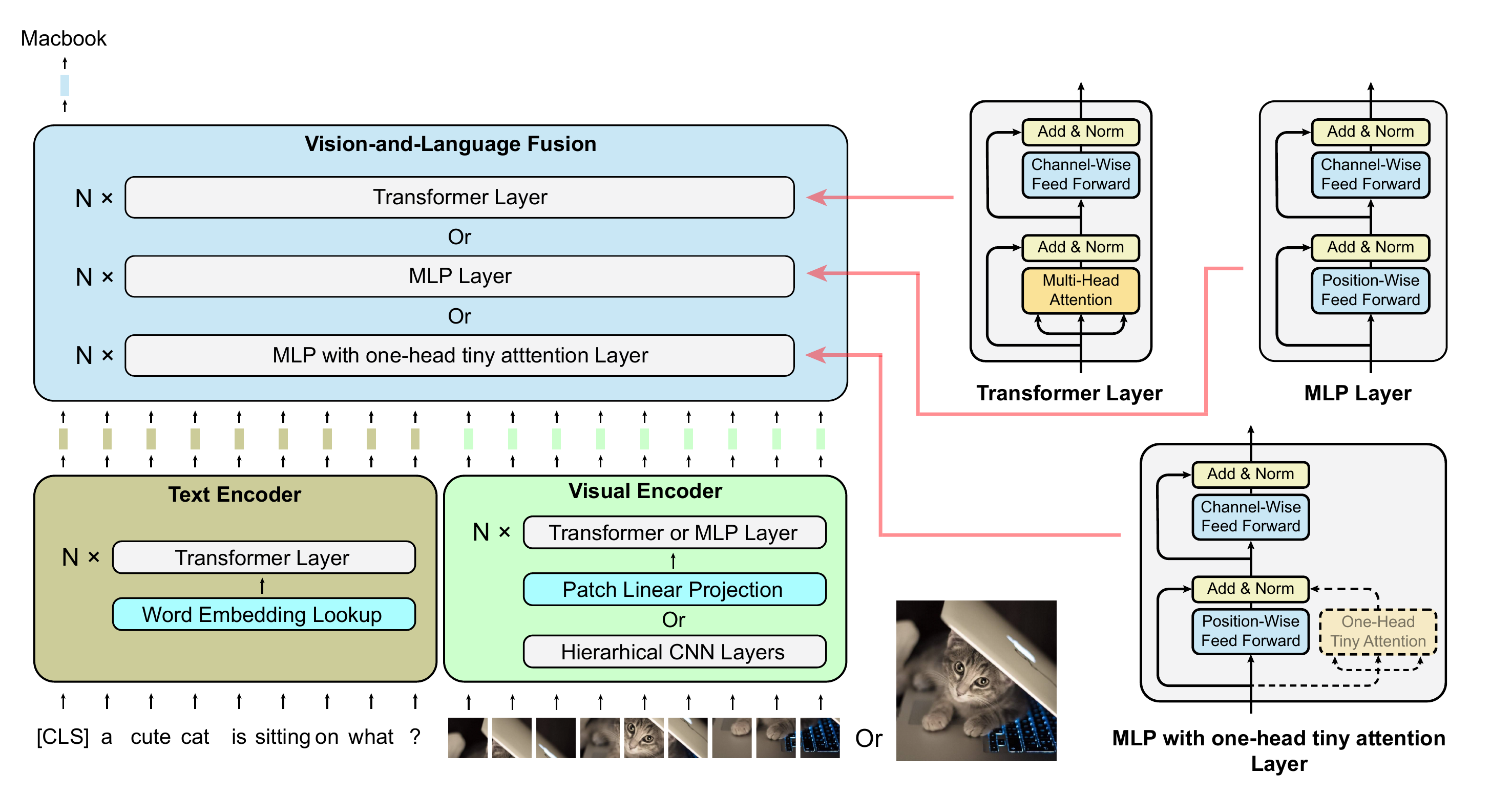}
    \vspace{-3mm}
\caption{\textbf{Overview of our framework for studying MLPs for vision-and-language (VL) modeling.} (Left) The backbone of our VL model; (Right) Three alternatives that we experimented for multimodal fusion, including transformer, MLP, and MLP with tiny attention.}
\label{fig:overview}
\end{figure*}

A natural but challenging question that comes in is: \emph{Can MLPs also replace transformers for vision-and-language (VL) modeling?} 
On the methodology side, it is natural to take advantage of the attention mechanism for cross-modal alignment, and it remains an open question that to what extent such inductive bias is essential to the performance on VL tasks.
This makes MLP architectures for VL modeling not only an essential missing piece for understanding the potential of using MLPs as a universal approximator, but also an insightful case study for learning cross-modal alignment without any inductive priors that used to be imposed by the attention mechanism.
On the practical side, if the answer is affirmative, an MLP architecture with identical MLP layers simplifies the efforts on accelerating and optimizing the computation of neural networks at various levels, potentially also providing parameter and FLOP savings.

To answer this question, we present \textsc{MLP-ViL}, an MLP model for vision and language, and initiate the first empirical study to explore how MLPs can be used for VL tasks.
As a first step, we replace transformers with MLPs for multimodal fusion, as shown in Figure~\ref{fig:overview} (the {\color{cyan}cyan} block), while keeping the original vision and text encoders intact, where
the vision encoder can be an object detector as used in most previous work~\cite{tan2019lxmert,lu2019vilbert,li2019visualbert,chen2020uniter,li2020oscar}, or a convolutional or transformer network that is advocated in more recent end-to-end VL methods~\cite{huang2020pixel,huang2021seeing,shen2021much,xue2021probing,li2021align}.
We conduct an extensive empirical study on MLPs for multimodal fusion, across 5 VL tasks and additional 5 robust VQA datasets,
and compare the performance of using transformer, MLP, and MLP with tiny one-head attention~\cite{liu2021pay}. Interestingly, we find that ($i$) without pre-training, using MLP for multimodal fusion has a noticeable performance gap with transformer; ($ii$) however, VL pre-training can help close the gap. Further, ($iii$) heavy multi-head attention is unnecessary, and adding tiny one-head attention to MLP is sufficient to retain most of the performance achieved by transformers.

These results are encouraging, and motivate us to ask an even \emph{bolder} question: \emph{can we have an all-MLP architecture for VL modeling, where the vision encoder is also replaced by pre-trained MLPs?} To answer this, as shown in Figure~\ref{fig:overview} (the {\color{brown}brown} and {\color{green}green} blocks), we only use a word embedding layer for text encoding, and leverage state-of-the-art vision MLPs (\emph{e.g.}, MLP-Mixer~\cite{tolstikhin2021mlp} and Permutator MLP~\cite{hou2021vision}) for image encoding, so that the entire model is composed of MLPs from end to end. We find that although, previous all-MLP architectures are claimed to perform as well as convolution networks and transformers on image recognition tasks, the performance of an all-MLP VL model is sub-optimal when compared to full-featured state-of-the-art VL models when both of them get pre-trained. However, a pre-trained all-MLP model is able to surpass a non-pre-trained transformer model, highlighting the importance of pre-training for all-MLP VL modeling.

To sum up our contributions:
\begin{itemize}[leftmargin=*]
\setlength\itemsep{-0.2em}
\item We initiate the first known effort to study MLPs for VL modeling, where the MLPs are first used for multimodal fusion only, then used for the whole model as well.
\item We conduct comprehensive experiments over 5 VL tasks, including VQA~\cite{antol2015vqa}, GQA~\cite{hudson2019gqa},  NLVR$^2$~\cite{suhr2019corpus}, visual entailment~\cite{xie2019visual}, and Flickr30k image-text retrieval to collect the evidence and limitation of using MLPs for vision and language modeling.

\item We perform analysis on 5 robust VQA benchmarks, including VQA-Rephrasings~\cite{shah2019cycle}, VQA-LOL~\cite{gokhale2020vqa}, Adversarial VQA (in domain~\cite{li2021adversarial} and out-of-domain~\cite{sheng2021human}), and GQA-OOD~\cite{kervadec2021roses}, and demonstrate that MLPs, despite of its parameterized nature that is subjected to over-fitting, with tiny attention can generalize to these robust benchmarks roughly to the similar degree as transformers.
\end{itemize}

\section{Related Work}
\paragraph{Vision-and-Language Pre-training.} 
Adopted from its original usage in NLP~\cite{kenton2019bert}, large-scale pre-training has prevailed in VL image-text modeling~\cite{li2019unicoder,li2020hero,lei2021less,li2021value,li2019visualbert,zhou2019unified}. 
Early approaches all used transformers for multi-modal fusion~\cite{tan2019lxmert,lu2019vilbert,chen2020uniter,su2019vl}, and were built on top of off-the-shelf object detection features~\cite{anderson2018bottom}. 
Additional efforts have been made in end-to-end pre-training that directly learn from raw image pixels~\cite{huang2020pixel,huang2021seeing} or image patches~\cite{kim2021vilt}. The recent CLIP-ViL~\cite{shen2021much} advanced VL pre-training by leveraging image encoder from CLIP~\cite{radford2021learning}, which is trained on a massive amount of web-crawled image-text pairs. 
Unlike prior efforts that all adopt transformers for multimodal fusion, we initiate the first empirical study of using MLPs for VL fusion without heavy reliance on inductive bias like self-attention as in transformers.

\paragraph{MLPs in Vision and NLP.}
Pioneered by the first three concurrent works on using an all-MLP architecture for image recognition~\cite{tolstikhin2021mlp, melas2021you, touvron2021resmlp}, there has been a surge of newly proposed MLP-like architectures for computer vision.
Though, the pioneer works build upon the idea of learning representations without inductive bias, which is realized by replacing self-attention with MLP in transformers, the descendants, like Vision 
Permutator~\cite{hou2021vision}, 
S$^2$-MLP~\cite{yu2021smlp}, and RaftMLP~\cite{tatsunami2021raftmlp}, improve model performance on ImageNet~\cite{krizhevsky2012imagenet} by re-injecting spatial-related priors. Moreover, rather than following the layer-by-layer paradigm as in transformers and MLP-Mixer that takes in flatten feature tokens at each layer, some later works, like AS-MLP~\cite{lian2021asmlp}, Hire-MLP~\cite{guo2021hire}, and CycleMLP~\cite{chen2021cyclemlp}, re-design the layer-wise feature representations and resort to CNN-like pyramid architectures. These more ``spatial-sensitive" designs have advanced performance of MLP-like architectures on image classification and other computer vision tasks. 
Besides computer vision, recently, gMLP~\cite{liu2021pay} and MLP singer~\cite{tae2021mlpsinger} have been successfully used for NLP and voice synthesis.
Unlike previous studies exploring MLPs for uni-modal data, we make the first attempt to use MLPs for multimodal modeling between vision and language.

\section{Method}

\subsection{The Backbone}
To carry out our study on MLPs for VL modeling, we use the same backbone design in all our experiments. As shown in Figure~\ref{fig:overview}, the backbone is composed of three components which can take various forms. The text encoder often consists of a word embedding layer and a stack of multiple layers of transformers on the top~\cite{kenton2019bert}. The vision encoder, however, can be in three different forms: ($i$) a patch linear projection layer that is applied to a list of evenly divided patches of the input image, and then a stack of homogeneous layers such as transformers~\cite{dosovitskiy2020image} or MLPs~\cite{tolstikhin2021mlp} on the top; ($ii$) a convolutional network~\cite{he2016deep} that takes in an input image and outputs a flatten list of grid features; and ($iii$) an object detector~\cite{anderson2018bottom} that takes in an input image and outputs a list of regional features (not shown in Figure~\ref{fig:overview}).
The text encoder and the vision encoder will encode the input question and image into two lists of feature vectors separately.
Then, the VL fusion module will take in all the feature vectors and produce the answer based on the \texttt{[CLS]} token. Following the recent state-of-the-art VL modeling paradigm~\cite{chen2020uniter, tan2019lxmert}, VL fusion is composed of a stack of $N$ identical layers which share the same input and output format. Our work focuses on comparing the effectiveness of MLPs and transformers for VL fusion, and we experiment on three different layers shown on the right of Figure~\ref{fig:overview}. In the follow subsections, we will describe the inner-workings of and the differences between the three layers.

\subsection{Transformer vs. MLP Layer}
\label{sec:diff_layer}
The well-established transformer layer is composed of ($i$) a multi-head attention block, which is meant to capture position-wise token interaction by aggregating information across tokens; and ($ii$) a channel-wise feed forward network (FFN), which is meant to refine the representation of each token individually. To facilitate direct comparison between transformers and MLPs, our MLP layer (as shown on the top right of Figure~\ref{fig:overview}) is modified from the transformer layer by replacing the multi-head attention block with a position-wise feed forward block.

To be specific, let $\mathbf{X}\in\mathbb{R}^{n \times d}$ be the token representations with sequence length $n$ and dimension $d$. The position and channel-wise feed forward blocks are defined as:
\begin{align}
    \mathbf{Y} &= \mathbf{P}  \sigma(\mathbf{Q} \mathbf{X}) \tag*{(Position-wise FFN)}\\
    \mathbf{Y} &= \sigma(\mathbf{X}\mathbf{U}) \mathbf{V} \tag*{(Channel-wise FFN)}
\end{align}
where $\sigma$ is an activation function such as GeLU~\cite{hendrycks2016gaussian}. $\mathbf{Q}\in\mathbb{R}^{h \times n}$ and $\mathbf{P}\in\mathbb{R}^{n \times h}$ define linear projections along the position dimension. $\mathbf{U}\in\mathbb{R}^{d \times h}$ and $\mathbf{V}\in\mathbb{R}^{h \times d}$ define linear projections along the channel dimension. $\mathbf{Y}\in\mathbb{R}^{n \times d}$ is the output which has the same dimension as $\mathbf{X}$. Note that the position-wise feed forward block is almost identical to the channel-wise feed forward block except that the forward is applied along the position dimension rather than the channel dimension. We intentionally keep the most simplified MLP form as in MLP-Mixer~\cite{tolstikhin2021mlp} in our experiments to study the potential of MLPs without any design priors. Optionally, following previous work~\cite{liu2021pay}, we also add a one-head tiny attention block (shown on the bottom right of Figure~\ref{fig:overview}) in the MLP architecture as the ablation study for the self-attention block. A detailed description regarding the difference between our one-head tiny attention and the multi-head attention are provided in the next subsection.

The existence of the non-linearity in the position-wise FFN ensures that input values will be able to implicitly affect the aggregation of the input tokens. Therefore, the position-wise FFN can be seen as an implicit attention layer over the input, where the attention weights are conditioned on each input token separately, similar to the Synthesizer~\cite{tay2021synthesizer}. This is different from the explicit self-attention in transformers, where token-token attention matrix depends on the input representation of both tokens.
The removal of non-linearity in position-wise FFN will result in a statically parameterized attention layer, where the attention weights will not condition on the input. In Section~\ref{sec:ablation_and_analysis}, we will compare different variants of the MLP layer.

The MLP layer is retrospective as it only contains linear projection and non-linearity, the most primitive form of neural networks, plus the transpose of input and output.\footnote{The position-wise FFN and channel-wise FFN are identical if the input and output of either one is transposed.} The layers in our model are also isotropic in that the input and output of each layer will have the same size, similar to transformers, or deep recurrent neural networks (RNNs). This design is different from pyramid-like CNNs, and can be easily enlarged by layer stacking.

\subsection{The Attention Mechanism}

\paragraph{Multi-Head Attention.}
Attention, as one of the key components in transformer, can be described as mapping a query and a set of key-value pairs to an output.
Typically, multi-head attention is used in transformers such that there will be $m$ parallel attention functions with different projection parameters for the queries, values and keys. The final output will be the concatenation of all the $m$ individual attention outputs. The network is formulated as:
\begin{align}
\mathbf{Q}^i=\mathbf{X}\mathbf{W_q}^i;\mathbf{K}^i=\mathbf{X}\mathbf{W_k}^i;\mathbf{V}^i=\mathbf{X}\mathbf{W_v}^i \\
\mathbf{Y}^i = \textit{Softmax}(\frac{\mathbf{Q}^i{\mathbf{K}^i}^T}{\sqrt{k}}) \mathbf{V}^i \label{equ:attention}\\
\mathbf{Y} = \textit{Concat}[\mathbf{Y}^0, \mathbf{Y}^1, \mathbf{Y}^2, ..., \mathbf{Y}^{m-1}]
\end{align}
where $\mathbf{X}\in\mathbb{R}^{n \times d}$ is the input. $\mathbf{W_q}^i\in\mathbb{R}^{d \times k}$, $\mathbf{W_k}^i\in\mathbb{R}^{d \times k}$, and $\mathbf{W_v}^i\in\mathbb{R}^{d \times h}$ are trainable parameters for each individual head. $\mathbf{Y}^i\in\mathbb{R}^{n \times h}$ is the output of individual attention function and $\mathbf{Y}\in\mathbb{R}^{n \times mh}$ is the final concatenated output. $\sqrt{k}$ is a scaling factor applied on the attention matrix. Oftentimes, $mh$ is set to be equal to $d$ such that the input $\mathbf{X}$ and output $\mathbf{Y}$ will have the same dimension.
The attention mechanism provides a robust inductive bias in which token-token interaction is based on the input representations of both tokens.
Multi-head attention allows the model to jointly attend to information from different representation
subspaces at different positions~\cite{vaswani2017attention}.

\paragraph{One-Head Tiny Attention.}
Following~\cite{liu2021pay}, to blend in and ablate the factor of attention in our MLP architecture, we optionally utilize a one-head tiny attention in parallel with the position-wise feed-forward network. The one-head tiny attention is formulated as:
\begin{align}
\mathbf{Q}=\mathbf{X}\mathbf{W_q};\mathbf{K}=\mathbf{X}\mathbf{W_k}\\
\mathbf{Y} = \textit{Softmax}(\frac{\mathbf{Q}\mathbf{K}^T}{\sqrt{k}}) \mathbf{X}
\end{align}
where $\mathbf{X}\in\mathbb{R}^{n \times d}$, $\mathbf{W_q}\in\mathbb{R}^{d \times k}$, $\mathbf{W_k}\in\mathbb{R}^{d \times k}$, and $\mathbf{Y}\in\mathbb{R}^{n \times h}$. Compared to the typical dot-product attention of transformers in Equation~\ref{equ:attention}, we further save the parameters by removing the value-projection and applying attention matrix directly on the input $\mathbf{X}$. 
The intermediate output of the MLP with one-head tiny attention layer is the normalized summation of the input, the output of position-wise FFN, and the output of the one-head tiny attention.

\begin{table}[t]
\tablestyle{5pt}{1.1} 
\def\w{20pt} 
\centering
    
    \begin{tabular}{lccc}
    \toprule
    \textbf{Model} & \textbf{\#Param.}  & \textbf{\#Param. in Attn.} & \textbf{FLOPs}\\
    \midrule
    Transformer & 75.6M & 18.9M (25.0\%) & 22.53G \\
    \textsc{MLP-ViL} & 60.2M & 0.0M (0.0\%) & 19.74G \\
    \textsc{MLP-ViL} + tiny att. & 61.0M & 0.8M (1.3\%) & 20.50G \\
    \bottomrule
    \end{tabular}
    \caption{\textbf{Key properties of different model variants for VL fusion}. The values within the bracket (in the third column) indicate the proportion of parameters allocated for attention.  
     }
    \label{tab:model_params}
\end{table}

\subsection{The Model Family}
\label{sec:overall_model}
Our main models for the VL fusion consist of a stack of 6 transformer layers, MLP layers, or MLP with tiny attention layers. To simplify the term, we name the model with 6 layers of MLP layers as ``\textsc{MLP-ViL}" and the one with tiny attention as ``\textsc{MLP-ViL} + tiny att.". To extract features from raw text and pixels, we use CLIP-ResNet (50$\times$4)~\cite{radford2021learning} and RoBERTa-Base~\cite{liu2019roberta} for vision and text encoders, respectively. In Section~\ref{sec:ablation_and_analysis}, we also examine the performance of the fusion modules with different vision and text encoders.
The hidden size of all the layers is set to 1024, and the hidden size of the tiny one-head attention is 64. Key properties of our architectures are summarized in Table~\ref{tab:model_params}. The number of parameters in \textsc{MLP-ViL} is 20\% less than that of a transformer model. The FLOPs of \textsc{MLP-ViL} is also 10\% less than that of transformers. More importantly, although, we add a tiny one head attention in the MLP layer, the proportion of actual parameters allocated for attention in such a model is only 1.3\%. The proportion is significantly less than that in transformers, where one-quarter of the parameters are used for attention. Despite of being light, we will show in Section~\ref{sec:main_fusion_results} that, the tiny one head attention can close the performance gap between MLP and transformer.

Finally, we also experiment with an all-MLP architecture for VL modeling. Identical to our main model, we use 6 MLP layers for VL fusion. We use Vision Permutator~\cite{hou2021vision} and Swin MLP~\cite{liu2021Swin} 
as the vision encoder\footnote{We choose them because they give better results on VQA than other MLP-like vision encoders. More results will be shown in Sec.~\ref{sec:ablation_and_analysis}.} and word embedding lookup layer as the text encoder. The resultant model is an attention-free and convolution-free architecture. To simplify the term, we name the above model ``all\textsc{MLP-ViL}".

\section{Experiments}
\subsection{Experimental Setup}
\paragraph{Pre-training Datasets.} We use image-text pairs in COCO~\cite{lin2014microsoft} and Visual Genome (VG)~\cite{krishna2017visual}, and visual question answering data from VQA~\cite{antol2015vqa, goyal2017making}, GQA~\cite{hudson2019gqa}, and VG-QA~\cite{zhu2016visual7w} for pre-training. We follow the Karpathy split~\cite{karpathy2015deep} and exclude 5k images for development and 5k image for testing from COCO. Table~\ref{tab:pretrain_datasets} gives a comparison between pre-training data in our experiment and other existing works. The total number of images is smaller than that of other existing works.

\begin{table}[t!]
\tablestyle{5pt}{1.1} 
\def\w{20pt} 
    \centering
    \begin{tabular}{llcc}
    \toprule
     \textbf{Method} & \textbf{Image Sources} & \textbf{\# of Image} & \textbf{\# of Image-Text}\\
    \hline
    UNITER & \scriptsize{COCO, VG, SBU, CC3M} & 4.2M & 9.6M \\
    VinVL & \specialcelll{\scriptsize{COCO, VG, SBU, CC3M,}\\ \scriptsize{Flickr30K, OpenImages}} &  5.7M & 8.9M$^\dagger$ (2.5M)\\
    ViLT &  \scriptsize{COCO, VG, SBU, CC3M} & 4.1M & 9.9M\\
    CLIP-ViL & \scriptsize{COCO, VG} & 180K & 9.2M\\
    \midrule
    Ours & \scriptsize{COCO, VG} & 170K & 8.7M (2.7M) \\
    \bottomrule
    \end{tabular}
\caption{\textbf{Pre-training dataset statistics}. The number in the bracket indicates VQA count. We strictly reserved 10K images from pre-training according to Karpathy split~\cite{karpathy2015deep}. ($\dagger$) the unique number of text-tag-image triples in VinVL.}
    \label{tab:pretrain_datasets}
\end{table}

\paragraph{Pre-training Objectives.} Following LXMERT~\cite{tan2019lxmert}, we consider 3 pre-training objectives: ($i$) Masked Language Modeling (MLM), we randomly mask out 15\% of words in the input sentence and train the model to reconstruct the masked words; ($ii$) Image-Text Matching (ITM), we randomly replace the aligned text with a different text with probability of 0.5, and train the model to classify whether the text corresponds to the image; ($iii$) Visual Question Answering (VQA), we use the annotated VQA data during pre-training and train the model to predict the correct answer given a question. More details are provided in Appendix.

\begin{table*}[t]
\tablestyle{5pt}{1.1} 
\def\w{20pt} 
    \centering
    \begin{tabular}{lccccccccc}
    \toprule
    \multirow{2}{*}{\textbf{Model}}
    & \multicolumn{2}{c}{\textbf{VQA}}
    & \multicolumn{2}{c}{\textbf{GQA}}
    & \multicolumn{2}{c}{\textbf{SNLI-VE}}
    & \multicolumn{2}{c}{\textbf{NLVR$^2$}}
    & \multirow{2}{*}{\textbf{Meta-Ave}}
    \\
      & test-dev & test-std & test-dev & test-std  & dev & test & dev & test-p \\
    \hline
    \hline
    \multicolumn{6}{l}{\textit{Previous models with pre-training}}\\
    \hline
    UNITER 
    & 72.70 & 72.91  & 59.99 & - & 78.59 &  78.28 & 77.18 & 77.85 & -\\
    VinVL 
    & 75.95 & 76.12 & \textbf{65.05} & \textbf{65.65}  & - & -  & \textbf{82.05} & \textbf{83.08} & - \\
    ViLT 
    & 71.26 & - & - & -  & - & -  &  75.70  & 76.13 & -\\
    CLIP-ViL 
    &  \textbf{76.48} & \textbf{76.70} & 61.42  & 62.93 & \textbf{80.61} & \textbf{80.20}  & - & - & -\\
    \hline
    \hline
    \multicolumn{6}{l}{\textit{Our models (different architectures for vision and language fusion) w/o pre-training}}\\
    \hline
    Transformer 
    &  74.04 & 74.22 & 58.58 & 58.65 & 78.44 & 78.69 & 54.03 & 52.30 & 66.11\\
    \textsc{MLP-ViL} 
    &  67.47 & 67.75 & 54.14 & 54.18 & 77.24 & 76.98 & 54.41 & 52.51 & 63.09\\
    \textsc{MLP-ViL} + tiny att. 
    &  73.80 & 73.93 & 59.35 & 59.68 & 78.32 & 78.30 & 53.90 & 51.90 & 66.14\\
    \hline
    \hline
    \multicolumn{6}{l}{\textit{Our models (different architectures for vision and language fusion) with pre-training}}\\
    \hline
    Transformer 
    & 75.42 (+1.38)& 75.74 & 61.33 (+2.75) & 61.90 & 79.64 & 79.90 & 79.55 & 78.36& \textbf{73.98}\\
    \textsc{MLP-ViL}  
    & 73.20 (+5.73) & 73.44 & 58.30 (+4.16) & 59.09 & 78.48 & 78.01 & 73.22 & 74.32 & 71.01\\
    \textsc{MLP-ViL} + tiny att. 
    & 74.77 (+0.97)& 74.93 & 61.28 (+1.93) &  61.90 & 79.50 &  79.32 & 76.23 & 77.92 & 73.23\\
    \hline
    \hline
    \multicolumn{6}{l}{\textit{All-MLP with pre-training (Permutator MLP)}}\\
    \hline
    all\textsc{MLP-ViL}  
    & 65.08 & 65.37 & 52.53 & 52.93 & 74.07 & 74.05 & 66.92 & 66.94 & 64.74 \\
    all\textsc{MLP-ViL} + tiny att. 
    & 67.65 & 67.86 & 55.83 & 55.41 & 75.03 & 75.13 & 72.72 & 71.36 & 67.62\\
    \hline
    \hline
    \multicolumn{10}{l}{\textit{All-MLP with pre-training (Swin MLP) }}\\
    \hline
    all\textsc{MLP-ViL} 
    & 66.38 & 66.56 & 54.59  & 54.99 & 74.49 & 74.44 & 68.17 & 70.42 & 66.26\\
    all\textsc{MLP-ViL}  + tiny att. 
    & 67.45 & 67.57 & 56.25 & 55.83 & 75.26  & 75.34 & 71.63 & 71.55 & 67.61\\
    \bottomrule
    \end{tabular}
    \caption{Results on visual reasoning tasks, including VQA, GQA, SNLI-VE, and NLVR$^2$. Training models from scratch on the NLVR$^2$ task only achieves accuracies comparable with random guess. \textbf{Meta-Ave} is the average score over all the evaluation sets.
    }
    \label{tab:visual_reasoning}
\end{table*}









\paragraph{Other Pre-training Details.} We train the model with a batch size of 512. We use AdamW as the optimizer. For better performance, the learning rate of VL fusion module is set to 5e-5, while the learning rate of the text and vision encoder are 1e-5.  The model is trained on 16 Nvidia V100 GPUs, and the pre-training takes around 5-6 days.

\paragraph{Downstream Tasks.}
We conduct a comprehensive evaluation of our models over a wide range of downstream tasks, including
VQAv2~\cite{goyal2017making}, GQA~\cite{hudson2019gqa}, Visual Entailment (SNLI-VE)~\cite{xie2018visual}, NLVR$^2$~\cite{suhr2019corpus}, and Image-Text Retrieval. In VQAv2~\cite{goyal2017making} and GQA~\cite{hudson2019gqa}, given an image and an input question, the model predicts an answer over a candidate pool. For NLVR$^2$~\cite{suhr2019corpus}, given a pair of images and a text description, the model judges the correctness of the description based on the visual clues in the image pair. For SNLI-VE~\cite{xie2018visual}, the model predicts whether a given image semantically entails a given sentence. 

\paragraph{Robust VQA Benchmarks.} Following ~\cite{li2020closer}, we also evaluate our model over on a suite of robust VQA benchmarks. VQA-Rephrasings~\cite{shah2019cycle} exposes VQA models to linguistic variations in questions, and measures consistency of model predicitions to different semantivally equivalent questions. VQA-LOL~\cite{gokhale2020vqa} examines the logical reasoning ability of a VQA model through questions containing logical compositions and linguistic transformations. Adversarial VQA~\cite{li2021adversarial, sheng2021human} datasets stress test a VQA model under adversarial attacks by human. GQA-OOD~\cite{kervadec2021roses} is based on out-of-distribution reorganization of the original GQA dataset, to evaluate question-oriented language bias in VQA models. More details can be found in Appendix.

\subsection{Results of MLPs for VL Fusion}
\label{sec:main_fusion_results}
The most important aspect of VL tasks lies in the modeling of VL fusion. Successful VL models~\cite{tan2019lxmert, chen2020uniter, li2019visualbert, lu2019vilbert} all rely on transformers for the fusion module. This is because the key building block of transformers, \emph{i.e.}, self-attention, brings the proper priors that can facilitate the modeling of complex VL interaction. This is proven by previous analysis~\cite{cao2020behind,hendricks2021decoupling}, all of which suggest that cross-modality attention between vision and language is crucial for the success of transformers on modeling VL tasks.
In this work, however, we would like to know whether MLP is able to perform VL fusion without explicitly designed attentions as in transformers. In what follows, we describe our findings.

\begin{table}[t]
\tablestyle{5pt}{1.1} 
\def\w{20pt} 
    \centering
    
    \resizebox{0.99\linewidth}{!}{
    \begin{tabular}{lcccccc}
    
    \toprule
    \multirow{2}{*}{\textbf{Model}}
    & \multicolumn{3}{c}{\textbf{Flickr30K (IR)}} & \multicolumn{3}{c}{\textbf{Flickr30K (TR)}}\\
      & R@1 & R@5 & R@10 & R@1 & R@5 & R@10 \\
    \hline
    \hline
    \multicolumn{7}{l}{\textit{Previous models with pre-training}}\\
    \hline
    ViLBERT & 58.2 & 84.9 & 91.5 & - & - & - \\
    Unicoder-VL & 71.5 & 91.2 & 95.2 & 86.2 & 96.3 & 99.0 \\
    UNITER & 72.5 &  92.4 & 96.1 &85.9 & 97.1 & 98.8 \\
    ImageBERT & - & - & - & - & - & - \\
    ViLT &  64.4 & 88.7 & 93.8 &  83.5 & 96.7 & 98.6 \\
    \hline
    \hline
    \multicolumn{7}{l}{\textit{Our models (different architectures for VL fusion) with pre-training}}\\
    \hline
    Transformer & \textbf{73.78} & 93.52 & 96.46 & \textbf{89.00} & 98.20 & 99.40 \\
    \textsc{MLP-ViL} &  65.42 & 91.20 & 95.26 & 78.90 & 97.50 & 99.20\\
    \textsc{MLP-ViL} + tiny att. & \textbf{73.78} & \textbf{93.76} & \textbf{97.06} & 88.11 & \textbf{98.40} & \textbf{ 99.50}\\
    \hline
    \hline
    \multicolumn{7}{l}{\textit{All-MLP with pre-training  (Permutator MLP)}}\\
    \hline
    all\textsc{MLP-ViL} & 42.68 & 75.08 & 83.82 & 55.74 & 85.61 & 91.91 \\
    all\textsc{MLP-ViL} + tiny att. & 57.40 & 85.40 & 91.00 & 72.62 & 92.16 & 95.93 \\
     \hline
    \hline
    \multicolumn{7}{l}{\textit{All-MLP with pre-training (Swin MLP) }}\\
    \hline
    all\textsc{MLP-ViL}  & 43.70  & 75.68 & 84.76 & 55.30 & 84.90 & 91.10 \\
    all\textsc{MLP-ViL} + tiny att.& 52.74 & 80.96  & 88.64 & 68.70 & 90.60 & 95.10\\
    \bottomrule
    \end{tabular}
    }
    \caption{Results on Flickr30k image-text retrieval. 
    }
    \label{tab:retrieval}
\end{table}

\paragraph{Without pre-training, using MLP for VL fusion has a noticeable performance gap when compared with transformers.} As shown in Table~\ref{tab:visual_reasoning}, when the models are trained from scratch using only the training data of VQA, GQA, and SNLI-VE, performance of \textsc{MLP-ViL} is $\sim$6.5 points lower on VQA, $\sim$4.5 points lower on GQA, and $\sim$2 points lower on SNLI-VE than those of transformers, indicating that the self-attention mechanism in the transformers indeed provides inductive bias beneficial for the performance.

\begin{table*}[t]
\tablestyle{5pt}{1.1} 
\def\w{20pt} 
    \centering
    \small
\resizebox{\textwidth}{!}{
    \begin{tabular}{lcccccccccccccc}
    \toprule
    \multirow{2}{*}{\textbf{Model}}
    & \multicolumn{5}{c}{\textbf{VQA-Rephrasings}}
    & \multicolumn{2}{c}{\textbf{VQA-LOL}}
    & \multicolumn{2}{c}{\textbf{Adversarial VQA}}
    & \multicolumn{4}{c}{\textbf{GQA-OOD}}\\
      & Acc. & CS(1)$\uparrow$ & CS(2)$\uparrow$ & CS(3)$\uparrow$ & CS(4)$\uparrow$ & Comp. & Supp. & In-domain & Out-of-domain & All  & Tail & Head & $\Delta\downarrow$\\
    \hline
    \hline
    \multicolumn{14}{l}{\textit{Previous models with pre-training}}\\
    \hline
    UNITER &  64.56 & 71.29 & 58.48 & 51.76 & 48.18 &  54.54 &  50.00 & 25.16 &  18.81 & 53.43 & 48.45 & 56.49 & 16.59 \\
    LXMERT & 67.20 & - & -  & - & - & 49.34 & 47.33 & - & 19.31 & 54.60 & 49.80 & 57.70 & 15.90 \\
    VILLA & 65.35 & 72.18 & 65.28 & 60.99 & 57.93 & 54.90 & 56.17 & 25.14 & 19.68 & 54.11 & 49.86 & 56.72 & 13.76\\
    MANGO & 65.80 & 72.66 & 66.03 & 61.92 & 58.95 &  \textbf{56.22} & \textbf{56.49} & - & -  & 54.47 & 50.24 & 57.07 & 13.59 \\
    \hline
    \hline
    \multicolumn{14}{l}{\textit{Our models (different architectures for vision and language fusion) w/o pre-training}}\\
    \hline
    Transformer & 66.46  & 73.58 & 67.66 & 63.91 & 61.16
    & 52.75 & 48.65 
    & 36.40 
    & \textbf{34.76} 
    & 53.22 & 49.81 & 55.47 & 11.36\\
    \textsc{MLP-ViL} & 59.48 &  66.47 & 59.89 & 55.99 & 53.23
    & 50.33 &  46.86
    & 30.90 
    & 26.36
    & 46.88 & 43.42 & 49.19 & 13.29\\
    \textsc{MLP-ViL} + tiny att. & 64.71 & 71.73 & 65.02 & 60.85 & 57.88
    & 51.55 &  49.10
    & 36.07 
    &  33.14  
    & 52.47 & 48.21 & 55.13 & 14.35 \\
    \hline
    \hline
    \multicolumn{14}{l}{\textit{Our models (different architectures for vision and language fusion) with pre-training}}\\
    \hline
    Transformer & \textbf{71.95} & \textbf{79.03} & \textbf{73.56} & \textbf{70.01} & \textbf{67.38}   
    & 54.14 & 52.44
    & \textbf{37.60} 
    & 33.79
    & \textbf{55.30}  & \textbf{51.50} & \textbf{57.78} & 12.19\ \\
    \textsc{MLP-ViL} & 69.09 & 76.09 & 70.30 & 66.67 & 64.02   
    & 53.91 & 52.98
    & 34.54 
    & 29.60
    &  51.43 & 47.37 & 54.09 & 14.19\ \\
    \textsc{MLP-ViL} + tiny att. & 70.37 & 77.43 & 71.81 & 68.23 & 65.59 
    &  53.42 &  51.23
    &  37.49 
    & 33.18
    & 53.72 & 51.32 & 55.36 & 7.87 \\
    \hline
    \hline
    \multicolumn{13}{l}{\textit{All-MLP with pre-training (Permutator MLP) }}\\
    \hline
    all\textsc{MLP-ViL} & 60.99 & 68.00 & 58.77 & 53.28 & 49.51   
    & 49.45 & 48.13
    & 29.99 
    & 23.39
    & 46.35 & 44.36 & 47.70 & \textbf{7.53}\\
    all\textsc{MLP-ViL} + tiny att.& 63.66 & 70.73 & 62.30 & 57.15 & 53.54   
    & 49.22 & 47.43
    & 30.92 
    &  24.57
    &  47.80 & 43.05 & 50.67 & 17.70\\
    \hline
    \hline
    \multicolumn{13}{l}{\textit{All-MLP with pre-training (Swin MLP) }}\\
    \hline
    all\textsc{MLP-ViL} & 60.09 & 67.18  & 58.34 & 53.10 & 49.51
    & 48.31 & 46.40
    & 29.52 
    &  22.98
    &  48.17  &  43.52 &  50.98 & 17.14 \\
    all\textsc{MLP-ViL} + tiny att.& 61.88 & 68.98  & 60.34 & 55.16 & 51.58
    & 49.37 & 49.81
    & 30.45 & 23.97
    & 47.85  & 42.67 & 51.04 & 19.62\\
    \bottomrule
    \end{tabular}
}
    \caption{Results on robust VQA benchmarks. We only compare our models with existing base-size models. 
    }
    \label{tab:robust}
\end{table*}

\paragraph{However, this gap can be reduced after VL pre-training.} For VQA, GQA and SNLI-VE, the performance boost brought by pre-training for \textsc{MLP-ViL} is larger than that for transformers. Specifically, as the numbers within the brackets in Table~\ref{tab:visual_reasoning} shown, pre-training can give a 5.74-points and 4.16-points improvement for VQA and GQA, which are noticeably larger than the improvement pre-training can give for transformers (1.38 for VQA and 2.75 for GQA). This suggests that immersing MLP in more data helps it to implicitly learn the generalizable inductive bias which otherwise was explicitly injected into the network design of transformers.

\paragraph{Adding tiny one-head attention to MLP is also able to substantially close this performance gap for the models both with and without pre-training.} Adding tiny attention, in spite of the fact that it only contributes 0.8M parameters (as explained in Table~\ref{tab:model_params}), is able to improve \textsc{MLP-ViL} in all the VL tasks, achieving performance very close to transformers on each individual task, and in the worst case only 0.81 points lower than transformers (VQA test-std).
Particularly, Table~\ref{tab:retrieval} shows that MLP with tiny attention can even surpass transformers on IR@5 and IR@10.

\subsection{Results of All-MLP Models}

\paragraph{All-MLP models are sub-optimal comparing to state-of-the-art full-featured models.}
In general, results of the all\textsc{MLP-ViL} are lower than that of full-featured VL models, ranging from $\sim$4 points drop in SNLI-VE to $\sim$7 points drop in VQA. This is expected, because ($i$) state-of-the-art VL models often involve built-ins such as pre-trained high-resolution object detection models~\cite{chen2020uniter, zhang2021vinvl}, language models~\cite{chen2020uniter, zhang2021vinvl}, or strong convolutional visual backbones~\cite{shen2021much}; ($ii$) state-of-the-art VL models use transformers that facilitate VL fusion. The results suggest that there is still room for improvement on using parameterized MLP for VL interaction modeling without reliance on inductive bias design.

\paragraph{However, it is surprising that all\textsc{MLP-ViL} can obtain reasonably decent results in spite of its unfavorable nature.} It is worth noting that our all-MLP (the second last row in Table~\ref{tab:visual_reasoning} and Table~\ref{tab:retrieval}) is able to obtain decent performance on VQA, SNLI-VE, and NLVR$^2$.
More importantly, the all$\textsc{MLP-VIL}$ can even achieve 66.26 on \textbf{Meta-Ave}, which is higher than that of a non-pre-trained transformer model (66.11, row 5 in Table~\ref{tab:visual_reasoning}, despite that transformers architecture is conceivable more expressive than vision and langauge modeling.
The finding would provide evidence for the potentials of an attention-and-convolutional-free MLP architecture as a general representation learner for VL interaction.

\subsection{Robustness Analysis}

The results in the previous sections provide promising evidence that MLP is able to learn how to effectively align visual and text features extracted from lower-level encoders without heavy reliance on self-attention.
However, since MLP is implicitly learning the alignments based purely on parameterized layers, we could run the risk of obtaining MLP layers that memorize surface patterns, resulting in representations that are neither generalizable to out-of-domain settings, nor robust to adversarial evaluation. Previous study found that MLP-Mixer is extremely vulnerable to universal adversarial perturbations~\cite{liu2021we}.
Therefore, we conduct additional experiments to compare the performance of MLP and transformer in a suite of robust VQA benchmarks.

As shown in Table~\ref{tab:robust}, we find that the general performance trend in previous sections still holds. More importantly, the performance gap between MLP and transformers is not widened when evaluated on those presumably ``harder" datasets. What is the most surprising is the delta value ($\Delta$) for GQA-OOD, which measures the relative performance gap on questions with most frequent answers (head) against those with least frequent answers (tail), all\textsc{MLP-ViL} model outperforms other model variations and previous VL pre-trained models by a large margin. The low delta value suggests that all\textsc{MLP-ViL} may suffer less from overfitting to the most frequent answers. 
The overall trend indicates that the degree of generalizability which MLP layers can be equipped for VL interaction modeling is similar to transformers.

\begin{figure}[t]
	\centering
    \includegraphics[clip,width=0.48\textwidth]{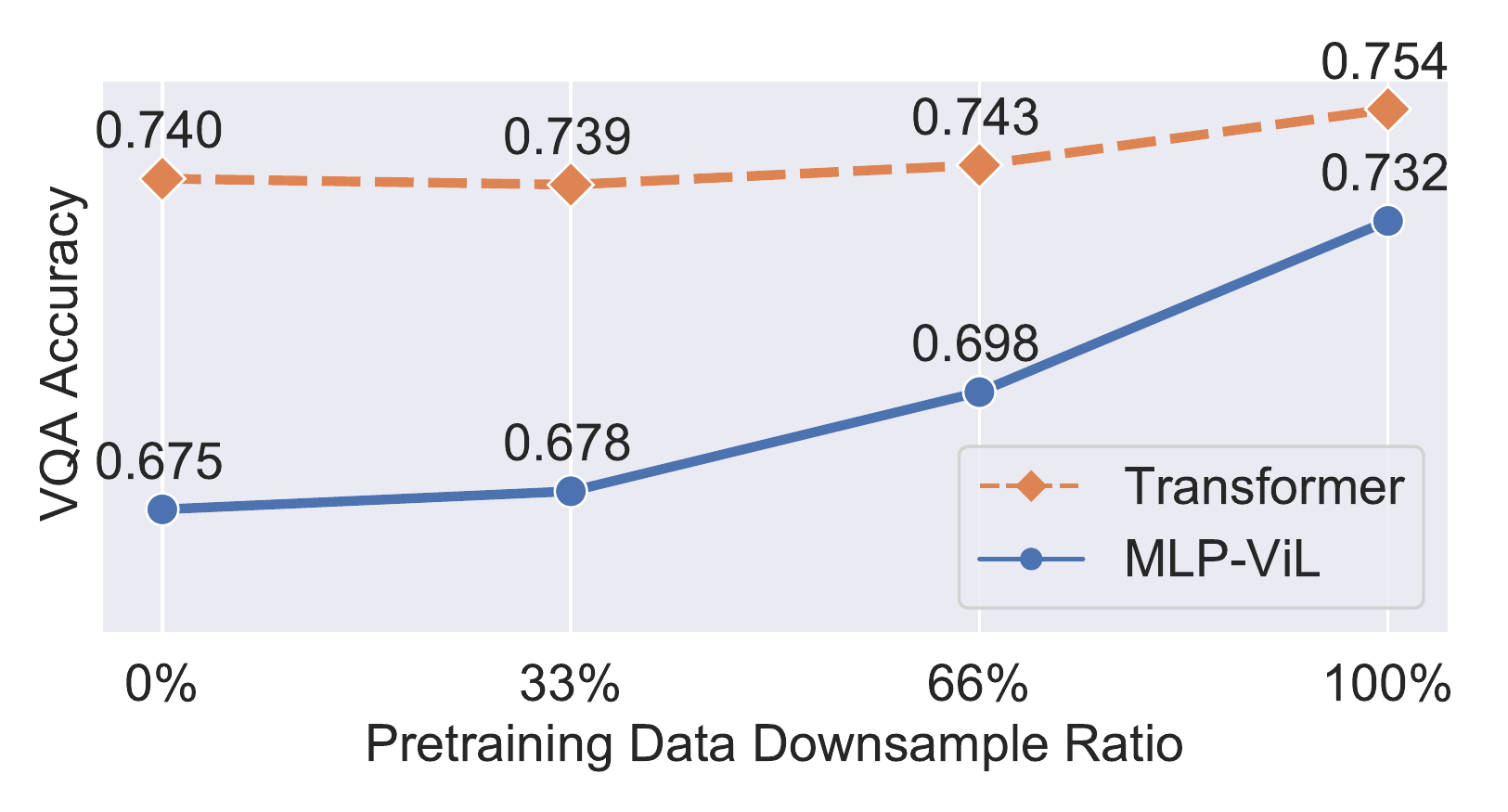}
\caption{Results of models pre-trained on downsampled data. \label{fig:downsample_training}}
\end{figure}

\section{Ablation and Analysis} \label{sec:ablation_and_analysis}
\subsection{Scaling Effect of Training Data}
In previous section, we show that large-scale pre-training can help close the gap between MLP and transformers for VL fusion. In order to understand how the size of the pre-training data can affect the model performance, we downsample the pre-training data to 33\% and 66\%, and then re-train the models and evaluate their performance on VQA. Results are shown in Figure~\ref{fig:downsample_training}. When immersed with large-scale data, MLP is able to scale up and got improved faster than transformers. This is similar to the finding in image recognition~\cite{tolstikhin2021mlp}, suggesting that MLP can achieve comparable results with transformers without designed priors if it can be scaled up. 

\subsection{Scaling Effect of Parameters}
\begin{figure*}[t!]
	\centering
    \includegraphics[clip,width=0.99\textwidth]{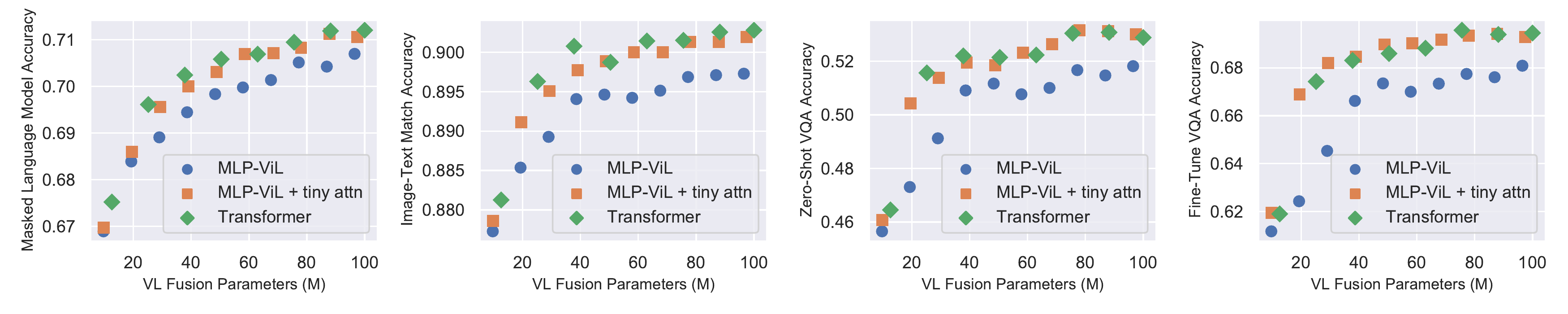}
\caption{Scaling effect of model parameters. Zero-Shot VQA Accuracy is the performance of pre-trained model without fine-tuning on VQA data only.}
\label{fig:efficiency_params}
\end{figure*}

We have studied the scaling effect of pre-training data for both transformer and \textsc{MLP-ViL}. We also investigate the scaling effect of parameters in VL fusion module by stacking one layer at a time for transformers, \textsc{MLP-ViL}, \textsc{MLP-ViL} + tiny att. Since pre-training all these models is extremely time-consuming, we use CLIP-ResNet (50$\times$4) as vision encoder and a word embedding look up layer, instead of RoBERTa-base,  as text encoder to make this scaling-up experiment affordable with limited computational resources. As shown in Figure~\ref{fig:efficiency_params}, both \textsc{MLP-ViL} and \textsc{MLP-ViL} + tiny att. can be scaled up by layer stacking. As expected, \textsc{MLP-ViL} + tiny att. and transformer are more parameter efficient than \textsc{MLP-ViL}. This may result from the difficulty in using purely parameterized \textsc{MLP-ViL} for cross-modal interaction. However, \textsc{MLP-ViL} is able to catch up to transformer with more parameters, suggesting the potential of MLP for VL modeling. More analysis about the experiments can be found in Appendix~\ref{app:parameters_and_flops}.

\begin{figure}[t]
	\centering
    \includegraphics[clip,width=0.48\textwidth]{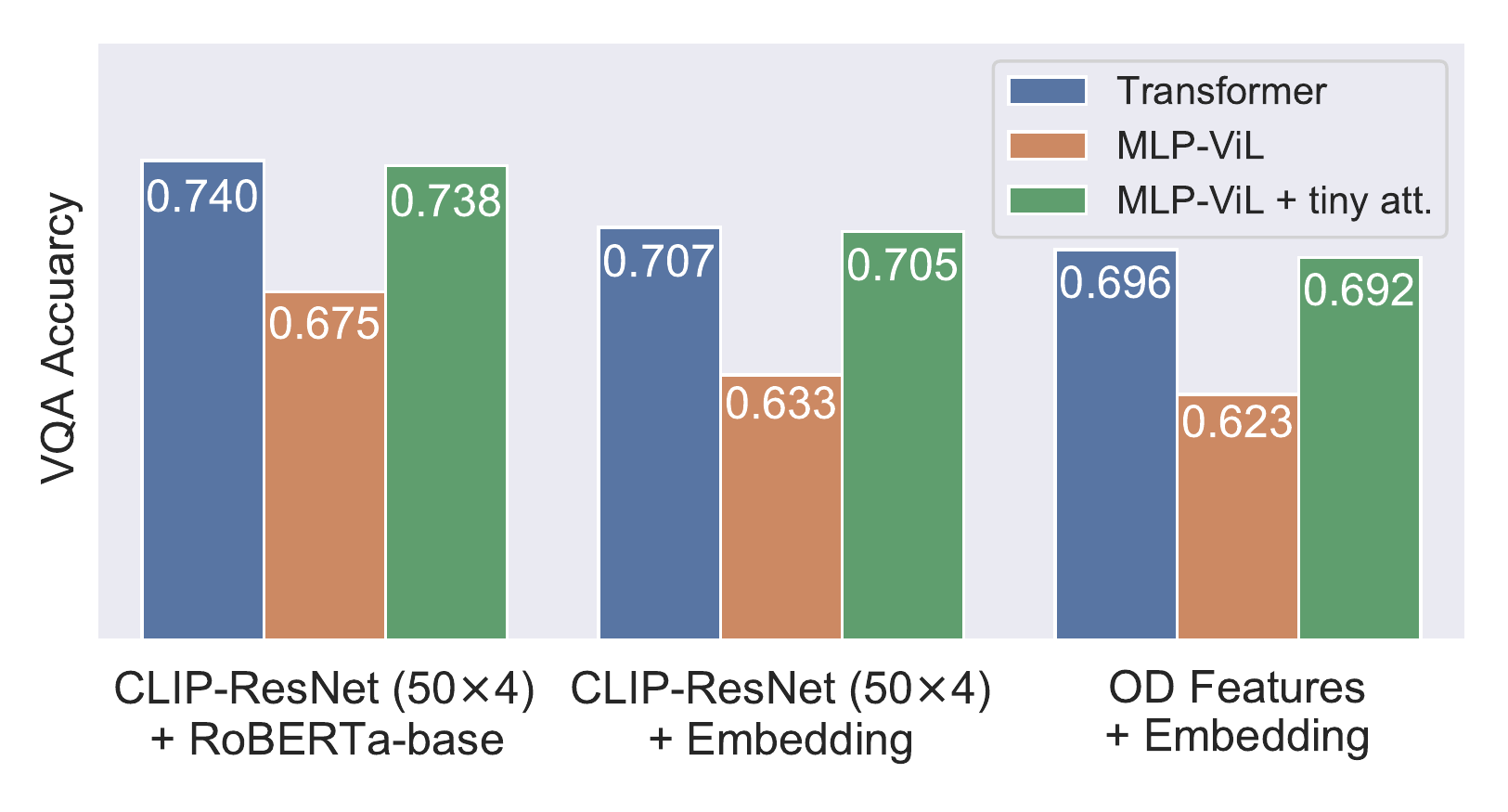}
\caption{Results of different models on VQA test-dev. The x-axis indicates different combination of vision and text encoder. The colors of the bar indicate different architecture for VL fusion. \label{fig:results_mlp_for_fusion}}
\end{figure}

\begin{table}[t!]
\tablestyle{5pt}{1.1} 
\def\w{20pt} 
\small
    \centering
    \begin{tabular}{lccc}
    \toprule
    \textbf{Vision Encoder} & \textbf{\#Param. } & \textbf{VQA} & \textbf{ImageNet-1k}\\
    \midrule
    Mixer-MLP & 229M & 60.9 & 84.2\\
    Permutator-MLP & 87M & 64.5 & 83.2\\
    Swin-MLP & 61M & 62.5 & 81.3\\
    Swin-Transformer & 228M & 69.9 & 87.3\\
    CLIP-ResNet (50$\times$4) & 178M & 70.1 & 83.5$^\dagger$\\
    \bottomrule
    \end{tabular}
    \caption{Results of using different vision encoders for VQA. ($\dagger$) indicates that the results for CLIP-ResNet (50$\times$4) on ImageNet-1k is in the zero-shot 5-top setting.}
    \label{tab:visual_encoder}
\end{table}

\begin{table}[t!]
\tablestyle{5pt}{1.1} 
\def\w{20pt} 
\small
    \centering
    \begin{tabular}{clc}
    \toprule
    & \textbf{Vision-and-Language Fusion} & \textbf{VQA}\\
    \midrule
    1 & Transformer & 74.04 (+6.57)\\
    2 & \textsc{MLP-ViL} + tiny att. & 73.80 (+6.33)\\
    3 & \textsc{MLP-ViL} & 67.47 (+0.00)\\
    4 & tiny attention & 72.59 (+5.12)\\
    5 & tiny attention $\times$ 2 & 72.54 (+5.07)\\
    6 & \textsc{MLP-ViL} (average pooling) & 66.08 (-1.39)\\
    7 & square MLP & 65.85 (-1.62)\\
    8 & square MLP + tiny att. & 72.41 (+4.93)\\
    \bottomrule
    \end{tabular}
    \caption{Results of variants of MLP layers for VL fusion. ``tiny attention" denotes replacing position-wise feed forward block with tiny one-head attention rather than adding them up. ``tiny attention $\times$ 2" denotes having 12 layers of tiny attention layer which doubles the number of layers in other experiments. ``average pooling" means that we use the average pooling of the last hidden output instead of the token representation of the \texttt{[CLS]} token as the output, following Mixer-MLP~\cite{tolstikhin2021mlp}. 
    ``square MLP" means that replacing $\mathbf{P}$ and $\mathbf{Q}$ in the position-wise FFN with 
    one square matrix $\mathbf{S}\in \mathbb{R}^{n\times n}$ such that $Y = \sigma(\mathbf{S} \mathbf{X})$ and thus the layer contains only one linear projection. The values in the bracket are the relative differences with the vanilla MLP performance.
    }
    \label{tab:mlp_variant}
\end{table}

\subsection{Effects of Vision \& Text Encoders} 
Other than the default setting explained in Sec.~\ref{sec:overall_model}, we also experiment with different combinations of vision and text encoders together with \textsc{MLP-ViL} and transformers without pre-training. As shown in Figure~\ref{fig:results_mlp_for_fusion}, despite the difference in the low level encoder, the statement that the performance gap between MLP and transformers can be closed by the tiny one-head attention holds in all scenarios, including using object detection features as vision encoder.
As all-MLP architectures emerged as a promising image recognition model, hinting the potentials of convolution and attention-free models, a question is raised that whether their performance can also be transferred to VL tasks like VQA. Therefore, we experimented with different MLP architectures for vision encoder and kept the text encoder and VL fusion module fixed. To focus the study on vision encoder, we used the word embedding as text encoder and 6 transformer layers as VL fusion module. We also used Swin transformer~\cite{liu2021Swin} which is the state-of-the-art on image recognition and the ResNet model from CLIP~\cite{radford2021learning} which is the state-of-the-art on zero-shot image recognition in the experiment as reference. As shown in Table~\ref{tab:visual_encoder}, the performance gap between MLP and the state-of-the-art models in VQA is larger than that in ImageNet-1k~\cite{krizhevsky2012imagenet}. Also, model performances on VQA are not aligned with the performances on ImageNet-1k. One comparison in particular that worth noting is that, although Mixer-MLP is able to achieve 84.2 on ImageNet-1k which is only 3.1 points below Swin transformer with the same amount of parameters, Mixer-MLP is 9 points below Swin transformer on VQA. The finding suggests that: ($i$) existing MLP vision encoders, despite their good performance on ImageNet-1k, are weaker at extracting vision features for VL modeling; and ($ii$) it is advisable to complement the general-purpose vision backbone evaluation with VL tasks.

\subsection{MLP Variants.} \label{sec:mlp_variants}
We also experimented with different variants of MLP layers on VQA with CLIP-ResNet (50$\times$4) and RoBERTa-base as vision and text encoders, respectively. Results are shown in Table~\ref{tab:mlp_variant}. We can see that, while replacing position-wise feed forward block with tiny one-head attention improves (row 3 and row 4), stacking more layers of tiny attention cannot further improve or even maintain the performance (row 4 and row 5). The position-wise feed forward block is also an irreplaceable component in the MLP with tiny attention layer (row 2 and row 4) that contributes to the performance. We also find that average pooling is slightly worse than \texttt{[CLS]} token pooling (row 3 and row 6) and that having two linear projections in the position-wise feed forward block is better than having only one square layer (row 3 and row 7).

\subsection{Weight Visualization}

\begin{figure}[t] 
	\centering
    \includegraphics[clip,width=0.48\textwidth]{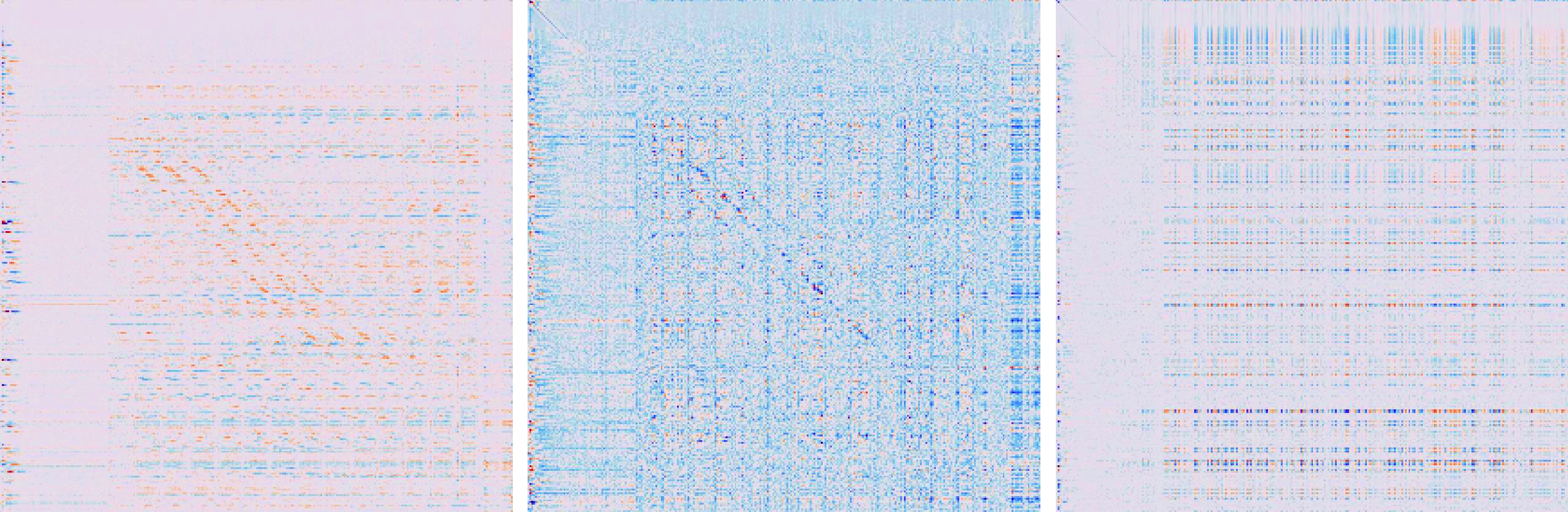}
\caption{Visualization of the simulated interaction matrices ($\mathbf{P}\mathbf{Q}$) in the first three position-wise FFN in \textsc{MLP-ViL}. The row and column start from text features and then vision features. Blue (Red) indicates low (high) value. \label{fig:vis}}
\end{figure}

As we can see in Figure~\ref{fig:vis}, there are apparent parameterized cross-modal interaction between feature inputs from vision and text encoder at the bottom MLP layer. As the layer goes up, the interaction becomes diffused and implicit, hinting that token representation at each position aggregates information from all the positions, including both vision and text, and becomes more intertwined. This encourages future interpretation analysis on decoding such parameterized cross-modal interaction.

\section{Conclusions \& Future Work}

We initiate the first empirical study on using MLPs for
VL modeling. Results reveal that although, there are noticeable performance gap between using MLPs and using transformers for VL fusion, large-scale pre-training and blending in tiny attention into MLPs can help close the gap. We further show that an all-MLP model can obtain reasonably decent performance when properly pre-trained and even surpass non-pre-trained transformers. Despite the improvement, MLP is, by design, less expressive than transformers in modeling complex cross-modal alignment.

Our findings shed light on the potential and limitations
of MLPs as general-purpose networks and encourage future
research on “decentralize” well-established transformers
for VL modeling. To further unfold the limit of learning
VL interaction without inductive design biases, future work could consider scaling up pre-training MLP architectures for VL both in term of data and model parameters if more computation resource is affordable.

\section*{Acknowledgement}
We sincerely thank Hao Tan, Zi-Yi Dou, Liunian Li, and Jie Lei for their helpful discussions. 
This work was partially supported by a Microsoft Investigator Fellowship, ARO Award W911NF2110220, and ONR Grant N000141812871. The views, opinions, and/or findings contained in this article are those of the authors and not of the funding agency.


{\small
\bibliographystyle{ieee_fullname}
\bibliography{egbib}
}

\appendix
\section*{Appendix}

\section{More Results}

\subsection{Result Discussion on all\textbf{\textsc{MLP-ViL}}}
We explore two vision encoder, Vision Permutator~\cite{hou2021vision} and Swin MLP~\cite{liu2021Swin}\footnote{\url{https://github.com/microsoft/Swin-Transformer}} for our all\textsc{MLP-ViL} models.
Note that the text encoder in all\textsc{MLP-ViL} is a simple text embedding lookup layer, and we adopt 6 layers of MLP layers for cross-modal fusion. The resultant all\textsc{MLP-ViL} is an attention-free and convolution-free architecture. 

Table~\ref{tab:visual_reasoning} presents the performance on visual reasoning tasks and Table~\ref{tab:robust} shows the evaluation on robust VQA benchmarks. Without reliance on inductive design bias,  we observe similar trends with Vision Permutator and Swin MLP for  all\textsc{MLP-ViL}, with a lower performance than full-featured VL models. When comparing different vision encoders for all\textsc{MLP-ViL},  Swin MLP shows stronger performance than Permutator MLP on standard visual reasoning tasks in Table~\ref{tab:visual_reasoning}, but worse performance on presumably more challenging datasets in Table~\ref{tab:robust}. The results suggest that although, existing all-MLP vision encoders (like Swin MLP and Permutator MLP) tend to improve their visual understanding ability by injecting spatial priors, their output representations can be distinct enough such that the advantage of one over the other is indecisive. In general, there is still much scope for improvement on all-MLP architectures both for vision encoder and for cross-modal fusion.

\subsection{Detailed Results on Image-Text Retrieval}
We further conduct image-text retrieval evaluation on both COCO~\cite{lin2014microsoft} and Flickr30k~\cite{plummer2015flickr30k} datasets under two settings: (1) zero-shot (Table~\ref{tab:supp_ret_zs}) and (2) finetuning (Table~\ref{tab:supp_ret_finetune}). We summarize our finding below.

First, there is a big gap between \textsc{MLP-ViL} and Transformer performance for zero-shot evaluation (15-20  on R@1). The performance gap can be reduced to 6-8 on R@1 after finetuning.  Second, adding tiny one-head attention to \textsc{MLP-ViL} is also able to substantially close this performance gap for zero-shot evaluation ($\sim$4 on R@1) or even enable stronger performance than Transformer on Flickr30K after finetuning. When considering all\textsc{MLP-ViL} architectures, they achieve sub-optimal performance, comparing to full-featured models. However, the added tiny one-head attention can still largely improve the model performance under both zero-shot and finetuning settings, sometimes achieving comparable performance to previous Transformer-based models (such as ViLBERT). These results further provide evidence for the potentials of MLP as a general representation learner for VL interaction.

\begin{table*}[t]
\tablestyle{5pt}{1.1} 
\def\w{20pt} 
    \centering
    \begin{tabular}{lcclccccccccc}
    \toprule
    \multirow{2}{*}{\textbf{Model}}
    & \multicolumn{3}{c}{\textbf{COCO (IR)}} & \multicolumn{3}{c}{\textbf{COCO (TR)}}
   & \multicolumn{3}{c}{\textbf{Flickr30K (IR)}} & \multicolumn{3}{c}{\textbf{Flickr30K (TR)}}\\
      & R@1 & R@5 & R@10 & R@1 & R@5 & R@10 & R@1 & R@5 & R@10 & R@1 & R@5 & R@10 \\
    \hline
    \hline
    \multicolumn{12}{l}{\textit{Previous models with pre-training}}\\
    \hline
    ViLBERT & - & - & - & - & - & - & 31.9 & 61.1 & 72.8 & - & - & - \\
    Unicoder-VL & - & - & - & - & - & - & 48.4 & 76.0 & 85.2 & 64.3 & 85.8 & 92.3\\
    UNITER & - & - & - & - & - & - & \textbf{66.2} & \textbf{88.4} & \textbf{92.9} & \textbf{80.7} & \textbf{95.7} & \textbf{98.0}\\
    ImageBERT & 32.3 & 59.0  & 70.2 & 44.0 & 71.2 & 80.4 &  54.3 & 79.6 & 87.5&  70.7 & 90.2 & 94.0\\
    ViLT &  \textbf{40.4} &  \textbf{70.0} &  \textbf{81.1} &  \textbf{56.5}&  \textbf{82.6} & \textbf{89.6}  &   55.0 &  82.5 & 89.8& 73.2 & 93.6 & 96.5\\
    \hline
    \hline
    \multicolumn{12}{l}{\textit{Our models (different architectures for vision and language fusion) with pre-training}}\\
    \hline
    Transformer & 37.28  & 65.34 & 76.55 & 53.28 & 79.60 & 87.54 &  56.30 & 81.86 & 89.18 & 73.90 & 91.60 & 96.20 \\
    \textsc{MLP-ViL} & 23.06 & 51.69 & 65.65 & 34.31 & 64.64 & 77.54&  38.68 & 70.30 & 80.50 &  53.00  & 79.60 & 88.30\\
    \textsc{MLP-ViL} + tiny att. & 33.15  & 61.15 & 72.86 & 48.86 & 75.92 & 84.76 &  52.40 & 79.12 & 86.86 & 69.20 & 88.10 & 94.20 \\
    \hline
    \hline
    \multicolumn{12}{l}{\textit{All-MLP with pre-training  (Permutator MLP)}}\\
    \hline
    all\textsc{MLP-ViL} & 13.33 & 38.74 & 53.65 & 18.08 & 46.22 & 60.72 & 15.62 & 40.42 & 54.92 & 22.88 & 48.65 & 61.94\\
    all\textsc{MLP-ViL}  + tiny att. & 24.69 & 53.64 & 67.25 & 34.04 & 64.46 & 75.98 & 31.68 & 61.06 & 72.28 & 43.00 & 71.50 & 80.30 \\
    \hline
    \hline
    \multicolumn{13}{l}{\textit{All-MLP with pre-training (Swin MLP) }}\\
    \hline
    all\textsc{MLP-ViL} & 14.99 & 40.58 & 56.31 & 20.76 & 48.64 & 62.66 & 17.68 & 43.06 & 57.06 & 24.90 & 54.80 & 68.50\\
    all\textsc{MLP-ViL} + tiny att.& 23.77 & 52.11 & 66.23 & 32.22 & 63.02 & 75.30 & 31.16 & 59.18 & 70.72 & 43.60 & 72.40 & 82.00\\
    \bottomrule
    \end{tabular}
    \caption{Results on zero-shot image-text retrieval. We only compare our models with existing base-size models.
    }
    \label{tab:supp_ret_zs}
\end{table*}

\begin{table*}[t]
\tablestyle{5pt}{1.1} 
\def\w{20pt} 
    \centering
    \begin{tabular}{lcclccccccccc}
    \toprule
    \multirow{2}{*}{\textbf{Model}}
    & \multicolumn{3}{c}{\textbf{COCO (IR)}} & \multicolumn{3}{c}{\textbf{COCO (TR)}}
   & \multicolumn{3}{c}{\textbf{Flickr30K (IR)}} & \multicolumn{3}{c}{\textbf{Flickr30K (TR)}}\\
      & R@1 & R@5 & R@10 & R@1 & R@5 & R@10 & R@1 & R@5 & R@10 & R@1 & R@5 & R@10 \\
    \hline
    \hline
    \multicolumn{12}{l}{\textit{Previous models with pre-training}}\\
    \hline
    ViLBERT & - & - & - & - & - & - & 58.2 & 84.9 & 91.5 & - & - & - \\
    Unicoder-VL & 48.4 &  76.7 & 85.9 & 62.3 &  87.1 & 92.8& 71.5 & 91.2 & 95.2 &  86.2 &  96.3 & 99.0\\
    UNITER &50.3 & 78.5 & 87.2 & 64.4 & 87.4 & 93.1 & 72.5 &  92.4 & 96.1 &85.9 & 97.1 & 98.8\\
    OSCAR & 54.0 & 80.8 & 88.5 &  70.0 &  91.1 & 95.5 & - & - & -& - & - & -\\
    VinVL & \textbf{58.1} & \textbf{83.2} & \textbf{90.1} & \textbf{74.6} & \textbf{92.6 }& \textbf{96.3}  & - & - & -& - & - & -\\
    ViLT &   42.7 &  72.9 & 83.1 & 61.5 & 86.3 & 92.7  &   64.4 & 88.7 & 93.8 &  83.5 & 96.7 & 98.6\\
    \hline
    \hline
    \multicolumn{13}{l}{\textit{Our models (different architectures for vision and language fusion) with pre-training}}\\
    \hline
     Transformer & 46.65 &  76.02 & 85.60 & 61.78 & 86.38  & 92.40 &  \textbf{73.78} & 93.52 & 96.46 & \textbf{89.00} & 98.20 & 99.40 \\
    \textsc{MLP-ViL} & 40.14 & 72.16 & 83.24 & 54.58& 82.56 & 90.36 &  65.42 & 91.20 & 95.26 & 78.90 & 97.50 & 99.20\\
    \textsc{MLP-ViL} + tiny att. & 44.98 & 75.19 & 85.11 &  60.60 & 86.10 &  92.42 & \textbf{73.78} & \textbf{93.76} & \textbf{97.06} & 88.11 & \textbf{98.40} & \textbf{ 99.50}\\
    \hline
    \hline
    \multicolumn{13}{l}{\textit{All-MLP with pre-training  (Permutator MLP)}}\\
    \hline
    all\textsc{MLP-ViL} & 22.92 & 53.19 & 67.62 & 31.29& 63.94 & 76.40 &  42.68 & 75.08 & 83.82 & 55.74 & 85.61 & 91.91 \\
    all\textsc{MLP-ViL} + tiny att. & 32.92 & 63.19 & 75.97 & 44.58 & 74.09 & 84.14 &  57.40 & 85.40 & 91.00 & 72.62 & 92.16 & 95.93 \\
    \hline
    \hline
    \multicolumn{13}{l}{\textit{All-MLP with pre-training (Swin MLP) }}\\
    \hline
    all\textsc{MLP-ViL} & 23.68 & 54.93 & 69.47 & 34.00 & 64.78 & 76.80 & 43.70  & 75.68 & 84.76 & 55.30 & 84.90 & 91.10 \\
    all\textsc{MLP-ViL} + tiny att.&  31.20 & 61.98 & 74.51 & 43.78 & 73.14 & 83.32 & 52.74 & 80.96  & 88.64 & 68.70 & 90.60 & 95.10\\
    \bottomrule
    \end{tabular}
    \caption{Results on image-to-text retrieval tasks after finetuning. We only compare our models with existing base-size models.
    }
    \label{tab:supp_ret_finetune}
\end{table*}

\subsection{Parameters and FLOPs}
\label{app:parameters_and_flops}
\begin{figure*}[t!]
	\centering
    \includegraphics[clip,width=0.99\textwidth]{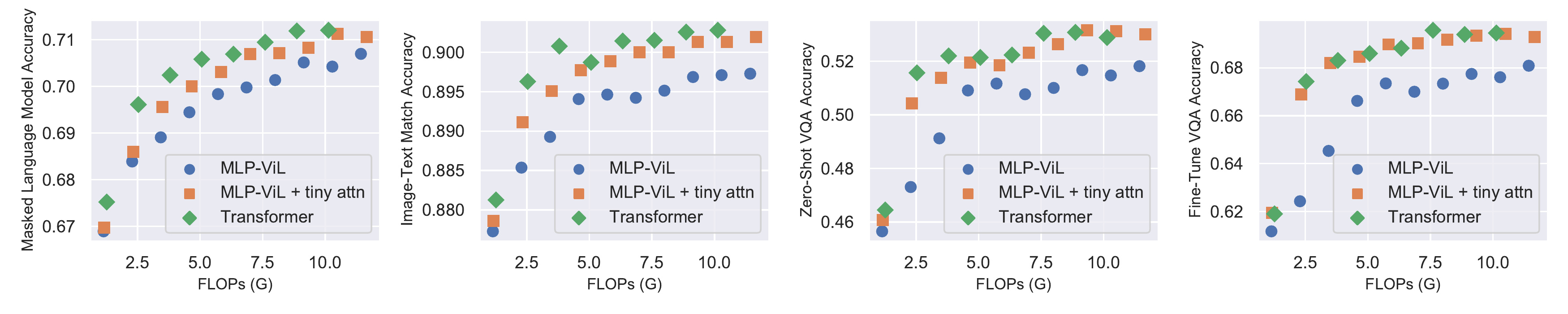}
    \vspace{-3mm}
\caption{FLOPs Comparsion. Zero-Shot VQA Accuracy is the performance of pre-trained model without fine-tuning on VQA data only.}
\label{fig:efficiency_flops}
\vspace{-3mm}
\end{figure*}

In the main paper, we demonstrate the scaling effect of both training data and parameters for \textsc{MLP-ViL} and Transformer.
Similarly, we show the FLOPs vs. model performance in Figure~\ref{fig:efficiency_flops} for this scaling-up experiment. All results illustrated in Figure~\ref{fig:efficiency_params} and~\ref{fig:efficiency_flops} are presented in Table~\ref{tab:efficiency_results}.

Table~\ref{tab:the_overall_models} summarizes the parameters, FLOPs and train/inference time of all model variants we studied in the main paper. 
To highlight the difference in the vision-and-language fusion module, we also show the parameters and FLOPs count only for the fusion module. To demonstrate the factor of attention, we show the parameters allocated for the attention mechanism. It is worth noting that our all\textsc{MLP-ViL} (Swin MLP) is faster and lighter than other models with less than one-fourth of the FLOPs and almost half of the parameters comparing to a full-featured transformer model.

\begin{table*}[t!]
\tablestyle{3pt}{1.1} 
\def\w{20pt} 
    \centering
    \begin{tabular}{lcccrrrr}
    \toprule
     \textbf{VL Fusion Model} & \textbf{\# of Layer} & \textbf{Parameter Size (M)} & \textbf{FLOPs (G)}&
     \textbf{MLM (Acc.)}&
     \textbf{ITM (Acc.)}&
     \textbf{ZS VQA (Acc.)}&
     \textbf{VQA (Acc.)}\\
    \midrule
Transformer                                   & 1  & 12.6 & 1.267 & 0.6752 & 0.8813 & 0.4645 & 0.6190 \\
Transformer                                   & 2  & 25.2 & 2.534 & 0.6961 & 0.8963 & 0.5157 & 0.6743 \\
Transformer                                   & 3  & 37.8 & 3.800 & 0.7024 & 0.9008 & 0.5220 & 0.6831 \\
Transformer                                   & 4  & 50.4 & 5.067 & 0.7058 & 0.8987 & 0.5214 & 0.6860 \\
Transformer                                   & 5  & 63   & 6.334 & 0.7069 & 0.9014 & 0.5223 & 0.6882 \\
Transformer                                   & 6  & 75.6 & 7.601 & 0.7094 & 0.9015 & 0.5304 & 0.6957 \\
Transformer                                   & 7  & 88.2 & 8.868 & 0.7119 & 0.9026 & 0.5308 & 0.6940 \\
Transformer                                   & 8  & 100  & 10.134 & 0.7120 & 0.9028 & 0.5288 & 0.6946 \\
\midrule
\textsc{MLP-ViL}            & 1  & 9.7  & 1.143 & 0.6689 & 0.8773 & 0.4565 & 0.6117 \\
\textsc{MLP-ViL}            & 2  & 19.3 & 2.286 & 0.6839 & 0.8854 & 0.4731 & 0.6243 \\
\textsc{MLP-ViL}            & 3  & 29   & 3.429 & 0.6891 & 0.8893 & 0.4912 & 0.6453 \\
\textsc{MLP-ViL}            & 4  & 38.6 & 4.572 & 0.6944 & 0.8941 & 0.5091 & 0.6662 \\
\textsc{MLP-ViL}            & 5  & 48.3 & 5.715 & 0.6983 & 0.8946 & 0.5117 & 0.6735 \\
\textsc{MLP-ViL}            & 6  & 57.9 & 6.857  & 0.6998 & 0.8942 & 0.5077 & 0.6700 \\
\textsc{MLP-ViL}            & 7  & 67.6 & 8.000 & 0.7013 & 0.8951 & 0.5100 & 0.6734 \\
\textsc{MLP-ViL}            & 8  & 77.2 & 9.143 & 0.7051 & 0.8969 & 0.5167 & 0.6775 \\
\textsc{MLP-ViL}             & 9  & 86.9 & 10.286 & 0.7042 & 0.8971 & 0.5147 & 0.6761 \\
\textsc{MLP-ViL}             & 10 & 96.5 & 11.429 & 0.7069 & 0.8973 & 0.5182 & 0.6809 \\
\midrule
\textsc{MLP-ViL} + tiny att. & 1  & 9.8  & 1.167 & 0.6698 & 0.8786 & 0.4607 & 0.6194 \\
\textsc{MLP-ViL} + tiny att. & 2  & 19.5 & 2.333  & 0.6860 & 0.8911 & 0.5043 & 0.6688 \\
\textsc{MLP-ViL} + tiny att. & 3  & 29.3 & 3.500 & 0.6956 & 0.8951 & 0.5138 & 0.6820 \\
\textsc{MLP-ViL} + tiny att. & 4  & 39   & 4.666 & 0.7000 & 0.8977 & 0.5195 & 0.6847 \\
\textsc{MLP-ViL} + tiny att. & 5  & 48.8 & 5.833 & 0.7031 & 0.8989 & 0.5184 & 0.6898 \\
\textsc{MLP-ViL} + tiny att. & 6  & 58.5 & 6.999 & 0.7069 & 0.9000 & 0.5231 & 0.6903 \\
\textsc{MLP-ViL} + tiny att. & 7  & 68.5 & 8.166  & 0.7071 & 0.9000 & 0.5264 & 0.6918 \\
\textsc{MLP-ViL} + tiny att. & 8  & 78   & 9.332 & 0.7082 & 0.9013 & 0.5316 & 0.6935 \\
\textsc{MLP-ViL} + tiny att. & 9  & 87.8 & 10.499 & 0.7113 & 0.9013 & 0.5312 & 0.6941 \\
\textsc{MLP-ViL} + tiny att. & 10 & 97.5 & 11.666 & 0.7106 & 0.9019 & 0.5300 & 0.6928\\
    \bottomrule
    \end{tabular}
    \vspace{-2mm}
\caption{Scaling up by layer stacking. For all the models in the table, we used CLIP-ResNet (50$\times$4) as vision encoder and word embedding look up layer as text encoder with different vision-and-language fusion modules.}
    \label{tab:efficiency_results}
    \vspace{-2mm}
\end{table*}
\begin{table*}[t]
\tablestyle{5pt}{1.1} 
\def\w{20pt} 
\centering
    \begin{tabular}{lcccc}
    \toprule
    \multirow{2}{*}{\textbf{Model}} & \textbf{\#Parameters}  & \textbf{\#Parameters in Attention} & \textbf{FLOPs}& \textbf{Time/100 steps (s)}\\
    \cmidrule{2-2}\cmidrule{3-3}\cmidrule{4-4}\cmidrule{5-5}
    & Fusion/Total & Fusion/Total & Fusion/Total & Train/Inference\\
    \hline
    \hline
    \multicolumn{4}{l}{\textit{Different architectures for multimodal fusion}}\\
    \hline
    Transformer & 75.6M/270.3M & 18.9M/40.1M & 22.53G/82.97G &  92.84/29.71\\
    \textsc{MLP-ViL} & 60.2M/254.9M & 0.0M/21.2M & 19.74G/80.18G & 90.48/29.15\\
    \textsc{MLP-ViL} + tiny attn & 61.0M/255.7M & 0.8M /22.0M & 20.50G/80.94G & 92.45/29.75\\
    \hline
    \hline
    \multicolumn{4}{l}{\textit{All-MLP (Permutator MLP)}}\\
    \hline
    all\textsc{MLP-ViL} & 60.6M/180.7M & 0.0M/0.0M & 21.89G/46.42G & 50.72/19.66\\
    all\textsc{MLP-ViL} + tiny attn & 61.4M/181.5M & 0.8M/0.8M & 22.79G/47.32G & 53.64/20.27\\
    \hline
    \hline
    \multicolumn{4}{l}{\textit{All-MLP (Swin MLP)}}\\
    \hline
    all\textsc{MLP-ViL} & 58.0M/151.1M & 0.0M/0.0M & 7.55G/18.05G & 32.24/10.66 \\
    all\textsc{MLP-ViL} + tiny attn & 58.8M/151.9M & 0.8M/0.8M & 7.71G/18.21G & 33.14/11.00\\
    \bottomrule
    \end{tabular}
    \vspace{-2mm}
    \caption{Properties of all model variants we studied. Training and inference time are profiled with batch size 32 on 1 Quadro
     RTX 8000 GPU. 
     }
    \label{tab:the_overall_models}
    \vspace{-3mm}
\end{table*}

\section{Experimental Setups for Downstream tasks}

In this section, we describe each downstream task and our implementation details. 

\subsection{Visual Reasoning Tasks}
\paragraph{Visual Question Answering (VQAv2)}\cite{antol2015vqa,goyal2017making} asks for
answers given pairs of an image and a question in natural
language. We follow the common practice to convert the open-ended question answering task to a classification task with a finite set of answer classes. We use the same answer vocabularies as the pre-training experiments, with 9,500 answers in total.  We fine-tune our models on the VQAv2 train and validation sets while reserving 5,000 validation images used in the Karpathy testing split~\cite{karpathy2015deep} and their related questions for internal validation. Binary cross entropy loss is used to supervise the model training.  We report both test-dev and test-std scores from the submission to the evaluation server. 

\paragraph{GQA}\cite{hudson2019gqa} is same as VQAv2 (\textit{i.e.}, answer related questions on a single image), but GQA requires more reasoning skills (e.g., spatial understanding and multistep inference). Similarly, we formulates GQA as a multi-class classification problem with a pre-defined answer vocabulary and use binary cross entropy loss to supervise the model training.  We follow~\cite{chen2021meta} to finetune our models on the balanced train and validation splits and report performance on test-dev and test-std splits from the submissions to the evaluation server.

\paragraph{Visual Entailment (SNLI-VE)}\cite{xie2019visual}  is a task derived from
Flickr30K images~\cite{plummer2015flickr30k} and Stanford Natural Language Inference (SNLI) dataset~\cite{bowman-etal-2015-large},
where the goal is to determine the logical relationship between a natural language statement and an image. Following UNITER~\cite{chen2020uniter}, we treat SNLI-VE as a three-way classification problem and finetune our models with
cross-entropy loss.

\paragraph{Natural Language for Visual Reasoning for Real (NLVR$^2$)}\cite{suhr2019corpus} is a binary classification task given triplets of two images and a question in natural language. We follow~\cite{li2020oscar,zhang2021vinvl,kim2021vilt} and adopt the pair method. Here, the
triplet input is reformulated into two pairs (question, image1) and (question, image2), and each pair goes through the model to obtain the joint image-text representations. A task-specific head, consisting of a two-layer MLP, takes the concatenation of both
representations as input and outputs the binary prediction. 
\subsection{Image-Text Retrieval}
We fine-tune our models on the Karpathy
split~\cite{karpathy2015deep} of COCO~\cite{lin2014microsoft} and Flickr30K~\cite{plummer2015flickr30k}. For image-to-text and text-to-image retrieval, we measure both zero-shot and fine-tuned performance. We follow the same setting of Image-text matching during pre-training to fineune our models for imate-text retrieval tasks. Specifically, We randomly replace the aligned image with a different image with the probability of 0.5. We adopt the single linear layer ITM head during pre-training to project the joint image-text representation to logits over binary class, and we compute negative log-likelihood loss to supervise the model training. Following~\cite{kim2021vilt}, we also add the word patch alignment (WPA) loss using
the inexact proximal point method for optimal transports
(IPOT)~\cite{xie2020fast}.

\subsection{Robust VQA Benchmarks}
\paragraph{VQA-Rephrasings}\cite{shah2019cycle} is based on VQAv2~\cite{goyal2017making}. It contains 3 human-provided rephrasings for 40K questions on 40K images from VQAv2 val split. In addition to accuracy, consistency in model predictions to different semantically equivalent questions is also used to measure the robustness of VQA models against linguistic variations. 

\paragraph{VQA-LOL}\cite{gokhale2020vqa} is introduced to examine the logical reasoning ability of a VQA model through questions containing logical compositions and linguistic transformations (negation, disjunction, conjunction, and antonyms).  It consists of two sub-datasets: VQA-LOL Compose (logical combinations of multiple closed binary questions about the same image in VQAv2) and VQA-LOL Supplement (logical combinations of additional questions based on external object and caption annotations about the images from COCO~\cite{lin2014microsoft}). Both sub-datasets share the same train/val images as VQAv2.

\paragraph{Adversarial VQA}\cite{li2021adversarial, sheng2021human} datasets are recently introduced to stress test a VQA model under adversarial attacks by human. It consists of two sub-datasets in-domain~\cite{sheng2021human} (based on original VQAv2 images from COCO) and out-of-domain~\cite{li2021adversarial} (based on diverse image sources other than COCO), collected with Human-and-Model-in-the-Loop~\cite{nie2019adversarial,kiela2021dynabench}.

\paragraph{GQA-OOD}\cite{kervadec2021roses} aims to examine the robustness of a VQA model in a setting where language priors cannot be relied upon for a correct prediction.  It is based on out-of-distribution reorganization of the original GQA dataset, features answer distribution shifts for both validation and testing, while keeps the same training split as GQA. 

We adopt the same strategy as VQA and use binary cross entropy to optimize the model training. For datasets built on the original VQAv2 val splits, we follow~\cite{li2020closer} to finetune all models on VQAv2 training split to avoid data contamination. For adversarial VQA datasets, we follow~\cite{li2021adversarial, sheng2021human} to evaluate models finetuned on VQAv2+VG-QA. The evaluation on GQA-OOD is conducted on models finetuned on GQA training split.

\subsection{Additional Implementation Details}
For all our main models, we use CLIP-ResNet (50$\times$4) as the vision encoder and adopt the same image prepossessing method as CLIP-ViL~\cite{shen2021much} with image resolution 480$\times$480. When using Swin MLP~\cite{liu2021Swin} and Vision Permutator~\cite{hou2021vision} as vision encoder, we follow the corresponding image prepossessing methods in their implementations with image resolution 224$\times$224.

\end{document}